\documentclass[conference]{IEEEtran}
\IEEEoverridecommandlockouts
\usepackage{cite}
\usepackage{amsmath,amssymb,amsfonts}
\usepackage{algorithmic}
\usepackage{graphicx}
\usepackage{textcomp}
\usepackage{xcolor}
\usepackage{eso-pic}
\usepackage{booktabs}  
\usepackage{colortbl} 
\usepackage{xcolor}   
\usepackage{multirow}  
\usepackage{graphicx} 
\usepackage{caption}  
\usepackage{amssymb}
\usepackage{booktabs}
\usepackage{amssymb}
\usepackage{makecell}
\usepackage{subcaption} 
\def\BibTeX{{\rm B\kern-.05em{\sc i\kern-.025em b}\kern-.08em
    T\kern-.1667em\lower.7ex\hbox{E}\kern-.125emX}}
\begin{document}

\title{HuM-Eval: A Coarse-to-Fine Framework for Human-Centric Video Evaluation}

\author{
\IEEEauthorblockN{Bingzi Zhang$^{1}$, Kaisi Guan$^{1}$, Ruihua Song$^{1\dagger}$}
\IEEEauthorblockA{$^{1}$Gaoling School of Artificial Intelligence, Renmin University of China\\
$^{\dagger}$Corresponding author}
}

\maketitle
\AddToShipoutPictureFG*{%
  \AtPageLowerLeft{%
    \hspace{0.7in}%
    \raisebox{0.25in}{%
      \parbox{6.8in}{%
        \scriptsize
        © 2026 IEEE. Personal use of this material is permitted.
        Permission from IEEE must be obtained for all other uses, in any current or future media,
        including reprinting/republishing this material for advertising or promotional purposes,
        creating new collective works, for resale or redistribution to servers or lists, or reuse
        of any copyrighted component of this work in other works.
      }%
    }%
  }%
}
\begin{abstract}

Video generation models have developed rapidly in recent years, where generating natural human motion plays a pivotal role.
However, accurately evaluating the quality of generated human motion video remains a significant challenge. Existing evaluation metrics primarily focus on global scene statistics, often overlooking fine-grained human details and consequently failing to align with human subjective preference.
To bridge this gap, we propose HuM-Eval, a novel human-centric evaluation framework that adopts a coarse-to-fine strategy. Specifically, our framework first utilizes a Vision Language Model to perform a coarse assessment of global video quality. It then proceeds to a fine-grained analysis, using 2D pose to verify anatomical correctness and 3D human motion to evaluate motion stability.
Extensive experiments demonstrate that HuM-Eval achieves an average human correlation of 58.2\%, outperforming state-of-the-art baselines.
Furthermore, we introduce HuM-Bench, a comprehensive benchmark comprising 1,000 diverse prompts, and conduct a detailed evaluation of existing text-to-video models, paving the way for next-generation human motion generation.
\end{abstract}


\section{Introduction}
\label{sec:intro}
Video generation models have achieved remarkable progress in recent years. Leading foundation models, such as Sora2~\cite{sora2024} and Veo3~\cite{Veo3}, can now produce videos with impressive semantic fidelity and complex scene dynamics. In this context, human-centric content stands out as the most pivotal category. Since humans are the primary subjects of visual storytelling, the ability to generate natural human motion is the key factor that determines a model's practical value. However, accurately evaluating the quality of generated human motion videos remains a significant challenge.
\begin{figure}[!ht]
    \centering
    \includegraphics[width=\linewidth]{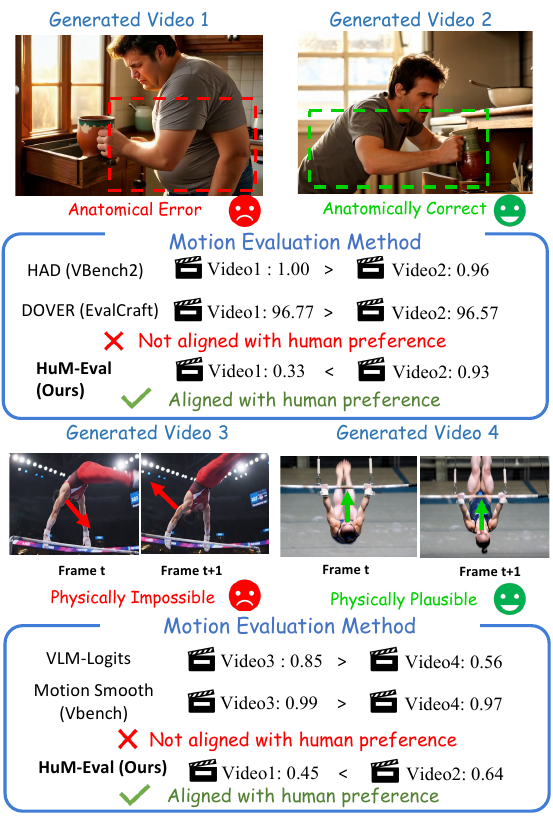} 
     \caption{Human video evaluation comparison. Baselines often fail to detect anatomical errors (top) and physically impossible motions like orientation flips (bottom). By leveraging explicit structural features, our metric correctly penalizes these artifacts, ensuring superior alignment with human perception.}
    \label{fig:teaser_right}
    \vspace{-2em}
\end{figure}
Existing evaluation methods generally fall into three categories. First, distribution-based metrics, such as FVD~\cite{unterthiner2018fvd}, measure the distance between generated and real data distributions. However, they cannot assess the quality of individual video samples. Second, rule-based metrics rely on measuring pixel-level frame changes, such as motion smoothness~\cite{huang2023vbench} or aesthetic scores~\cite{ling2025vmbench}. Third, Vision Language Model (VLM)-based metrics like VideoScore~\cite{videoscore} or VideoAlign~\cite{videoalign} leverage VLMs fine-tuned on human annotations to judge general quality. Crucially, both rule-based and VLM-based approaches prioritize global scene quality over specific human details. Consequently, they fail to capture critical artifacts, such as anatomical distortions and motion instability.

To address this, we propose HuM-Eval, a coarse-to-fine framework inspired by human perception. Our method integrates three key components to assess human motion video quality comprehensively. First, we employ a Perceptual Visual Prior Score using a VLM to evaluate the coarse-grained visual quality. Then, to capture fine-grained details, we introduce explicit physical constraints: the Anatomical Structure Score, which utilizes 2D pose confidence to detect structural distortions and errors, and the Motion Stability Score, which leverages 3D motion analysis to penalize physical instability. This combination ensures our metric rigorously covers both global scene quality and local human fidelity.

Building on this framework, we further introduce HuM-Bench, a comprehensive benchmark covering three core categories: Body-Motion Only, Human-Object Interaction, and Human-Human Interaction. Constructed via a progressive ``Word-Phrase-Prompt" pipeline, HuM-Bench comprises 1,045 high-quality prompts. We evaluate eight existing video generation models on our benchmark. To ensure a reliable ground truth, we conducted a rigorous human study, providing high-quality labels for Anatomical Correctness and Motion Smoothness.

Extensive experiments demonstrate that our metric achieves superior alignment with human preference compared to existing baselines. Furthermore, we provide a detailed performance analysis of recent video generation models, offering insights into their current capabilities. We hope this work serves as a meaningful reference to facilitate future research in human motion video generation.

\begin{figure*}[t] 
    \centering
    
    \includegraphics[width=0.95\linewidth]{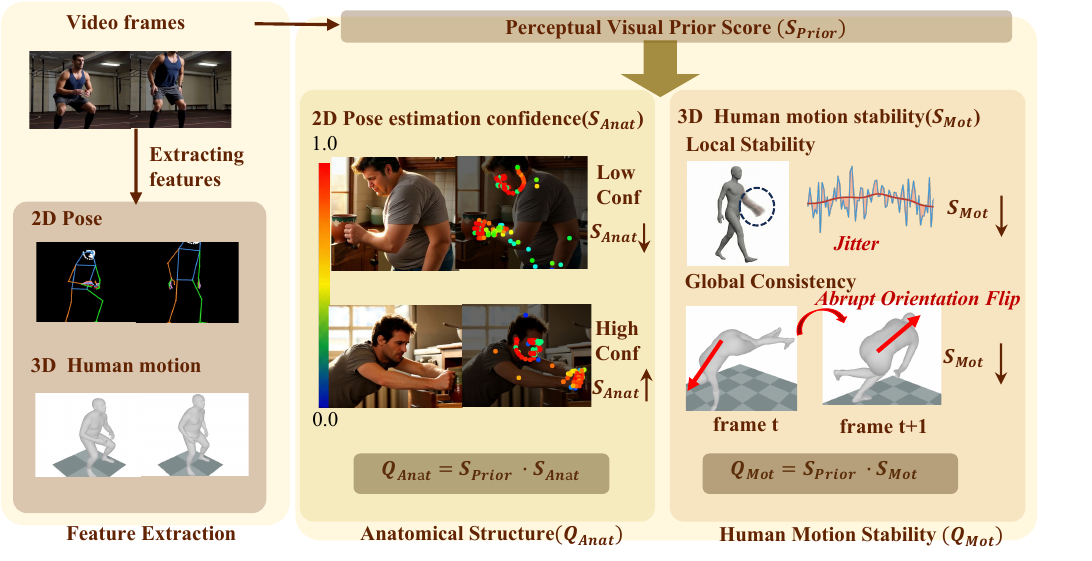} 
    
    \caption{The Framework Overview. Our proposed Coarse-to-Fine evaluation strategy. The VLM  provides a holistic perceptual ($S_{Prior}$), which is subsequently refined by explicit 2D pose features (left) and 3D motion constraints (right) to yield the final quality scores.}
    \label{fig:framework} 
\end{figure*}

\section{Related Work}
\subsection{Video Generation Models}
The field of video generation has witnessed a paradigm shift from UNet-based architecture~\cite{3dunet} to Diffusion Transformers~\cite{Peebles2022DiT} architecture. Video Generation Models such as Sora~\cite{sora2024} have achieved impressive results in synthesizing high-fidelity video content~\cite{wan2025wan,kong2024hunyuanvideo} and aligning with user instructions~\cite{guan2025etva}. More recently, advanced models like Veo3~\cite{Veo3} have demonstrated the ability to generate video with synchronized audio~\cite{wang2025jointdit,guan2025bridgedit}. However, despite these successes, generating realistic human-centric motion remains an essential challenge. Models frequently fail to maintain correct body structure~\cite{ling2025vmbench} and realistic motion~\cite{zhao2025motionphysical}. Addressing this limitation is a critical step towards building a comprehensive world model.

\subsection{Automatic Evaluation on Generated Human Motion}
Current evaluation methods for human motion in video generation can be broadly categorized into three types: 
% Distribution-based
(1) Distribution-based methods: Methods such as FVD~\cite{unterthiner2018fvd}, and FVMD~\cite{liu2024fvmd} treat generated videos as a probability distribution, measuring the feature distance between the generated videos and real videos. 
%   Rule-based
(2) Rule-based methods: These approaches rely on predefined rules to evaluate motion smoothness by quantifying temporal variations between frames. Existing works include calculating optical flow bias, interpolation changes~\cite{huang2023vbench}, or aesthetic score differences~\cite{ling2025vmbench}\, using these inter-frame variations as indicators of motion quality.
% MLLM-based
(3) VLM-based methods: Methods like VideoScore~\cite{he2024videoscore} and VideoAlign~\cite{liu2025videoalign}
utilize Vision Language Models (VLMs) fine-tuned on human annotations. These models excel at evaluating overall video quality, achieving high consistency with human preference.
However, these metrics operate at a scene level and solely on temporal continuity between pixels, ignoring that a motion can be smooth yet anatomically incorrect. This paper focuses on fixing these issues.

\section{HuM-Eval Framework}
\subsection{Overview}

In our pilot user study on evaluating generated human-centric videos, we find that human subjects are primarily sensitive to the overall visual quality, and then to the errors in human appearances (e.g., twisted legs) and physically implausible movement (e.g., abrupt orientation flip). However, existing works often overlook fine-grained details of human motion.
To address this limitation, we propose a human-centric evaluation framework that adopts a coarse-to-fine strategy. As shown in Fig~\ref{fig:framework}, the framework contains three parts.
First, we introduce a Perceptual Visual Prior score $S_{Prior}$ using a VLM. This part provides a coarse evaluation of video quality, serving as a prior score for the subsequent fine-grained evaluation.
Second, we propose the Anatomical Structure Score $S_{Anat}$ to evaluate human anatomical structure quality in generated videos.
Third, we propose the Motion Stability Score $S_{Mot}$ to evaluate the human motion stability in the generated videos. Finally we propose integrating them to reflect the true user experiences.

\subsection{Perceptual Visual Prior Score}
\label{sec:vlm_metric}
When humans evaluate a video, they intuitively judge the overall quality before examining details. To simulate this process, we introduce the Perceptual Visual Prior Score. 

Specifically, given a generated video $\mathcal{V}$, we employ a pretrained VLM (here is Qwen3-VL-8B-Instruct~\cite{qwen3vl}) to query its general visual quality.
We design a prompt $\mathcal{P}$ that provides comprehensive assessment criteria (covering clarity, stability, aesthetics, and realism) to ground the model's judgment. The prompt concludes by asking:
\begin{quote}
\textit{``...Is this a high-quality and usable video? Respond ONLY with `Yes' or `No'.''}
\end{quote}

 We then extract the logits corresponding to the positive token $\mathcal{T}^+ =$ ``Yes'' and the negative token $\mathcal{T}^- =$ ``No''.
The Perceptual Visual Prior Score $S_{Prior}$ is calculated by applying a softmax operation over these logits:
\begin{equation}
\setlength{\abovedisplayskip}{3pt} 
\setlength{\belowdisplayskip}{3pt}
    S_{Prior} = \frac{\exp(\mathcal{L}(\mathcal{T}^+))}{\exp(\mathcal{L}(\mathcal{T}^+)) + \exp(\mathcal{L}(\mathcal{T}^-))}
\end{equation}
where $\mathcal{L}(\cdot)$ denotes the logit value output by the VLM. 

This score represents the holistic perception of the video, acting as a coarse-grained score. To effectively capture fine-grained details—such as anatomical distortions and motion jitter, we subsequently introduce explicit features to refine this initial assessment.

\subsection{Evaluation of Anatomical Structure Quality}
Shifting the focus to fine-grained human details, we first quantify the human anatomical structure quality $Q_{Anat}$.

Our core premise is that a high-quality generated human should be not only visually clear but also anatomically plausible. So we introduce the Anatomical Structure Score $S_{Anat}$ to refine $S_{Prior}$.
To measure this, we utilize Sapiens-Pose (2B)~\cite{khirodkar2024sapiens}, a pose estimator with a 133-keypoint topology covering the body, hands, and face. We leverage the insight that this high-capacity model, trained on massive real-world data, effectively acts as a structural discriminator. 

We utilize the keypoint estimation confidence as a proxy for $S_{Anat}$. High confidence indicates a clear, anatomically valid structure recognized by the pre-trained model, while low confidence signals potential artifacts such as blurred joints or twisted limbs that deviate from natural human topology.

To quantify $S_{Anat}$, we calculate the average confidence scores strictly for visible body parts. A body part is considered valid only if the average confidence of its keypoints exceeds the threshold $\tau=0.3$ (we adopt this threshold directly from the official Sapiens visualization protocol). For each frame $t$, the frame score $s_t$ is calculated by averaging the confidence of all keypoints belonging to these visible parts.

Formally, the final $S_{Anat}$ is computed by averaging the frame-level scores across the entire video sequence:
\begin{equation}
\label{eq:anat_score}
    S_{Anat} = \frac{1}{T} \sum_{t=1}^{T} s_t
\end{equation}
where $T$ denotes the total number of frames and $s_t$ represents the average confidence of visible body parts in frame $t$.

Finally, to utilize $S_{Anat}$ as a refinement coefficient, we normalize it against real-world distributions. We sample videos from the ActivityNet dataset~\cite{caba2015activitynet} to determine the empirical bounds, calculating the maximum ($max_{real}$) and minimum ($min_{real}$) $S_{Anat}$ from real footage. The raw score is then min-max normalized to $\hat{S}_{Anat} \in [0, 1]$. The final Anatomical Structure Quality is derived by modulating the coarse prior:
\begin{equation}
    Q_{Anat} = S_{Prior} \cdot \hat{S}_{Anat}
\end{equation}
This ensures the final score reflects holistic realism constrained by anatomically valid bounds derived from real-world data.
\subsection{Evaluation of Human Motion Stability}

While the coarse prior $S_{Prior}$ captures holistic visual perception, it  lacks depth information and exhibits limited sensitivity to temporal dynamics. Consequently, it often fails to detect physically impossible movements (e.g., sudden depth jumps). To address this, we introduce the human motion stability score ($S_{Mot}$), employing GVHMR~\cite{gvhmr} to lift the motion analysis into 3D space for fine-grained refinement.
We decompose human motion stability into two distinct aspects: Local Stability ($\mathcal{S}_{local}$), which targets the high-frequency jitter of joints relative to the body-centric frame, and Global Consistency($\mathcal{S}_{global}$), which evaluates the stability of human orientation within the world coordinate system.
To quantify high-frequency artifacts (e.g., jitter), we propose the $\mathcal{S}_{local}$ based on 3D angular Jerk. We distinguish unnatural noise from valid fast motion by calculating the deviation of the raw jerk from its Gaussian-smoothed trend. The score is derived by averaging this deviation magnitude across all frames:
\begin{equation}
\setlength{\abovedisplayskip}{3pt}
\setlength{\belowdisplayskip}{3pt}
    \mathcal{S}_{local} = \phi \left( \frac{1}{T} \sum_{t=1}^{T} \left\| \dddot{\Theta}_t - \mathcal{G}(\dddot{\Theta}_t) \right\|_2 \right)
\end{equation}
where $\dddot{\Theta}_t$ denotes the frame-level jerk, $\mathcal{G}(\cdot)$ represents a temporal Gaussian filter, and $\phi(\cdot)$ is a mapping function that normalizes the value into a perceptual score.

While local metrics handle joint jitter, we further introduce the $\mathcal{S}_{global}$ to penalize holistic orientation anomalies (e.g., unnatural flipping or snap turns).
To evaluate $\mathcal{S}_{global}$, we detect unnatural orientation flips by tracking two projected vectors: the body's vertical axis $\mathbf{v}^{up}_t$ and horizontal heading $\mathbf{v}^{head}_t$. We calculate the angular deviations of these vectors between adjacent frames to identify sudden flips. 
\begin{equation}
\setlength{\abovedisplayskip}{3pt}
\setlength{\belowdisplayskip}{3pt}
    \mathcal{S}_{global} = \phi \left( \max_{t} \left\{ 1 - \mathbf{v}^{up}_t \cdot \mathbf{v}^{up}_{t+1}, \ 1 - \mathbf{v}^{head}_t \cdot \mathbf{v}^{head}_{t+1} \right\} \right)
\end{equation}
This formulation ensures that any abrupt change in the human's global orientation—whether a sudden fall or a discontinuous turn—is effectively captured.

Analogous to the normalization protocol for $S_{Anat}$, we calibrate these raw kinematic scores using real-world priors. We utilize the Motion-x++ dataset~\cite{motionx} to calculate the empirical bounds for both $\mathcal{S}_{local}$ and $\mathcal{S}_{global}$. The raw metrics are then normalized to the unit interval $[0, 1]$. 
We first consolidate the local and global normalized constraints into a unified Motion Stability Score $S_{Mot}$:
\begin{equation}
    S_{Mot} = \hat{\mathcal{S}}_{local} \cdot \hat{\mathcal{S}}_{global}
\end{equation}
The final quality score $Q_{mot}$ is then derived by utilizing $S_{Mot}$ as a multiplicative penalty to modulate the perceptual prior. This gating mechanism ensures that any kinematic instability effectively suppresses the coarse visual score:
\begin{equation}
    Q_{Mot} = S_{Prior} \cdot S_{Mot}
\end{equation} 

\section{Our Proposed HuM-Bench}

\subsection{Our videos categories and generation}
We classify human motion into three categories: (1) Body-Motion Only (BMO), which focuses on human movements in free space including Simple Actions (e.g., walking, waving) and Skill-based Motions (e.g., gymnastics, martial arts); (2) Human-Object Interaction (HOI), involving tasks where humans interact with external items (e.g., playing a guitar, riding a bicycle); and (3) Human-Human Interaction (HHI), which covers interaction between multiple people (e.g., shaking hands, dancing in pairs).

To construct the benchmark, we employ a progressive ``Word-Phrase-Prompt" pipeline using LLMs. Starting with action verbs from datasets Inter-x~\cite{xu2024inter} and Motion-x++~\cite{motionx}, we expand them into structured phrases that define specific motions. These phrases are finally enriched into 1,045 prompts, incorporating stylistic and environmental contexts.

\subsection{Evaluation Subset and Human Annotation}
We sample a representative subset of 100 prompts for human annotation. To benchmark existing state-of-the-art video generation models, we evaluate five open-source models (Wan2.2-14B~\cite{wan2025wan},  SANA-Video(2B)~\cite{chen2025sana}, Open-Sora 1.2~\cite{lin2024opensora-plan}, HunyuanVideo~\cite{kong2024hunyuanvideo}, CogVideoX-1.5(5B)~\cite{yang2024cogvideox} ) and three closed-source models (Veo3, Seedance, and Hailuo).

We recruit 10 professional annotators to independently rate each video on a 5-point Likert scale across two key dimensions: \textit{Anatomical Correctness} and \textit{Motion Smoothness}. For the collected ratings, we perform an inter-rater consistency analysis and select the scores from the seven most consistent annotators to compute the mean score. For videos with significant variance in ratings, two additional expert reviewers conduct to calibrate the scores. 

\section{Experiments}
\subsection{Experimental Setup}

\noindent\textbf{Datasets.} 
We employed the previously constructed evaluation subset and manual annotation to benchmark 8 representative video generation models.

\noindent\textbf{Evaluation Metrics.} We employ Spearman's $\rho$ to evaluate the alignment between automated scores and human ratings. 

\noindent\textbf{Baseline Selection.} 
To prove the effectiveness of our HuM-Eval framework, we select representative metrics from the Rule-based and VLM-based methods.

\begin{itemize}
    \item \textbf{Rule-based metrics}: 
    (i) DOVER score~\cite{wu2022disentangling}, Flow score, and Warping Error from EvalCrafter~\cite{evalcrafter}; 
    (ii) Motion Smoothness, Time Flickering, and Human Anatomy from VBench/VBench2~\cite{huang2023vbench,vbench2}; 
    (iii) Motion Smoothness Score (MSS), Object Integrity Score (OIS), and Temporal Coherence Score (TCS) from VMBench~\cite{ling2025vmbench}.

    \item \textbf{VLM-based metrics}: 
    (i) Visual Quality, Temporal Consistency, and Factual Consistency from VideoScore~\cite{he2024videoscore}; 
    (ii) Motion Quality and Visual Quality from VideoAlign~\cite{liu2025videoalign}; 
    (iii) For Qwen3-VL-8B~\cite{qwen3vl}, we utilize two strategies: prompting for direct score and for logits probability.
\end{itemize}

\begin{table}[htbp]
\centering
\caption{Comparison of different evaluation methods using Spearman's $\rho$, higher indicates better alignment with human ratings. $ACS$ and $MSS$ denote Anatomical Correctness scores and Motion Smoothness scores rated by expert annotators. \textbf{Bold} indicates the best performance, and \underline{underline} indicates the second best.}
\label{tab:main_results}
\begin{tabular}{@{}lcc@{}} 
\toprule
\multirow{2}{*}{\textbf{Method}} & \multicolumn{2}{c}{\textbf{Spearman's $\rho$} ($\uparrow$)} \\ 
 & \textbf{$ACS$} & \textbf{$MSS$} \\ 

% Rule-based 
\rowcolor{gray!20} \multicolumn{3}{@{}l}{\textit{\textbf{Rule-based Methods}}} \\ 
EvalCrafter\cite{evalcrafter} (DOVER aesthetic) & 0.449 & 0.459 \\
EvalCrafter\cite{evalcrafter} (DOVER technical) & 0.437 & 0.456 \\
EvalCrafter\cite{evalcrafter} (Flow-score) & 0.323 & 0.295 \\
EvalCrafter\cite{evalcrafter} (Warping Error) & 0.308 & 0.338 \\
VBench\cite{huang2023vbench} (Motion Smoothness) & 0.273 & 0.298 \\
VBench\cite{huang2023vbench} (Time Flickering) & 0.054 & 0.080 \\
VBench2\cite{vbench2} (Human Anatomy) & 0.054 & 0.080 \\
VMBench\cite{ling2025vmbench} (MSS) & 0.503 & 0.495 \\
VMBench\cite{ling2025vmbench} (OIS) & 0.094 & 0.116 \\
VMBench\cite{ling2025vmbench} (TCS) & 0.292 & 0.280 \\ \midrule

\rowcolor{gray!20} \multicolumn{3}{@{}l}{\textit{\textbf{VLM-based Methods}}} \\ 
VideoScore\cite{videoscore} (Visual Quality) & 0.019 & -0.011 \\
VideoScore\cite{videoscore} (Temporal Consistency) & 0.112 & 0.084 \\
VideoScore\cite{videoscore} (Factual Consistency) & 0.055 & 0.028 \\
MQ (VideoAlign) \cite{videoalign} & 0.350 & 0.373 \\
VQ (VideoAlign) \cite{videoalign} & 0.309 & 0.335 \\
Qwen3-VL-8B (direct scoring)\cite{qwen3vl} & 0.277 & 0.286 \\
Qwen3-VL-8B (logits-based)\cite{qwen3vl} & \underline{0.565} & \underline{0.556} \\ \midrule

% Ours
\textbf{HuM-Eval (Ours)} & \textbf{0.593} & \textbf{0.572} \\ 
\bottomrule
\end{tabular}
\end{table}

\subsection{Alignment with human annotation}
Table \ref{tab:main_results} reports the Spearman's $\rho$ between automated metrics and human ratings. Our proposed framework performs better than the baselines, achieving the highest correlations in both Anatomical Correctness $ACS$ and  Motion Smoothness $MSS$.
Specifically, our results significantly surpass the best rule-based metric (\textit{MSS}).
Furthermore, our method outperforms the leading VLM approach (\textit{Qwen3-VL-8B (logits)}) across both dimensions.
These results suggest that while VLMs provide strong general perception, integrating explicit features for structural integrity and motion stability can be beneficial for accurate human-centric video evaluation.

\subsection{Ablation Study}
\label{sec:ablation}
In Table~\ref{tab:ablation_side_by_side}, the VLM-based perceptual prior $S_{Prior}$ achieves a respectable baseline correlation (0.556 for $MSS$ and 0.565 for $ACS$), indicating its capability to provide a reliable coarse assessment of overall visual quality. Integrating fine-grained physical scores ($S_{Anat}$ and $S_{Mot}$) as a refinement step further improves the alignment across both dimensions, showing that the addition of physical features enhances the performance, validating the effectiveness of our coarse-to-fine strategy.

\begin{table}[htbp]
\centering
\small
\renewcommand{\arraystretch}{1.2}
\caption{Ablation studies on  (Left) human
 motion stability and (Right) anatomical structure quality.}
\label{tab:ablation_side_by_side}

\setlength{\tabcolsep}{5pt}

\begin{tabular}{@{}ccc@{}}
\toprule
\multicolumn{2}{c}{\textbf{Components}} & \textbf{$\rho$} \\ 
\cmidrule(r){1-2} \cmidrule(l){3-3} 
\textbf{$S_{Prior}$} & \textbf{$S_{Mot}$} & \textbf{$MSS$} \\ \midrule
 & \checkmark & 0.475 \\
\checkmark & & 0.556 \\
\checkmark & \checkmark & \textbf{0.571} \\ \bottomrule
\end{tabular}
\quad \quad 
\begin{tabular}{@{}ccc@{}}
\toprule
\multicolumn{2}{c}{\textbf{Components}} & \textbf{$\rho$} \\ 
\cmidrule(r){1-2} \cmidrule(l){3-3} 
\textbf{$S_{Prior}$} & \textbf{$S_{Anat}$} & \textbf{$ACS$} \\ \midrule
 & \checkmark & 0.562 \\
\checkmark & & 0.565 \\
\checkmark & \checkmark & \textbf{0.593} \\ \bottomrule
\end{tabular}

\end{table}

\begin{table}[t]
    \centering
    \caption{Quantitative Evaluation on HuM-Bench. We report the quality scores for $Q_{Anat}$ and  $Q_{Mot}$ across representative T2V models. }
    \label{tab:leaderboard}
    
    \footnotesize 
    \renewcommand{\arraystretch}{1.2} 
    \setlength{\tabcolsep}{14pt} 
    
    \begin{tabular}{lcc}
        \toprule
        \textbf{Model} & \textbf{$Q_{Anat}$} & \textbf{$Q_{Mot}$} \\
        \midrule

        \rowcolor{gray!15}
        \multicolumn{3}{l}{\textit{\textbf{Closed-source Models}}} \\
        Veo & 0.723 & \underline{0.855} \\
        Hailuo & 0.712 & 0.835 \\
        Seedance & \underline{0.766} & 0.840 \\
        \midrule
        
        \rowcolor{gray!15}
        \multicolumn{3}{l}{\textit{\textbf{Open-source Models}}} \\
        Wan2.2 & \textbf{0.783} & \textbf{0.891} \\
        HunyuanVideo & 0.556 & 0.774 \\
        SANA & 0.759 & 0.849 \\
        CogVideoX & 0.172 & 0.247 \\
        OpenSora & 0.042 & 0.056 \\
        
        \bottomrule
    \end{tabular}
    \vspace{-1em}
\end{table}

\subsection{Benchmark Results}
\subsubsection{Quantitative Evaluation Results}

Table~\ref{tab:leaderboard} presents the quantitative comparison of 8 representative T2V models on our human-centric benchmark. The results indicate a progressive improvement in model capabilities regarding human generation.
Wan2.2-14B achieves the highest scores in both anatomical structure quality (0.783) and human motion stability (0.891), showing a strong capability in maintaining anatomical correctness and motion smoothness. Other high-performing models, including Seedance, SANA-Video(2B), and Veo3, also exhibit competitive results, with scores generally exceeding 0.70 in both dimensions. The results also reveal a discernible performance gap between different generations of models. While recent models demonstrate substantial improvements in structural integrity, earlier open-source versions such as CogVideoX-1.5(5B) and OpenSora-1.2 show limited capability in human-centric generation, particularly in motion smoothness, where scores remain below 0.30.

\subsubsection{Performance Across Motion Categories}
\begin{figure}[t]
    \centering
    \includegraphics[width=0.9\linewidth]{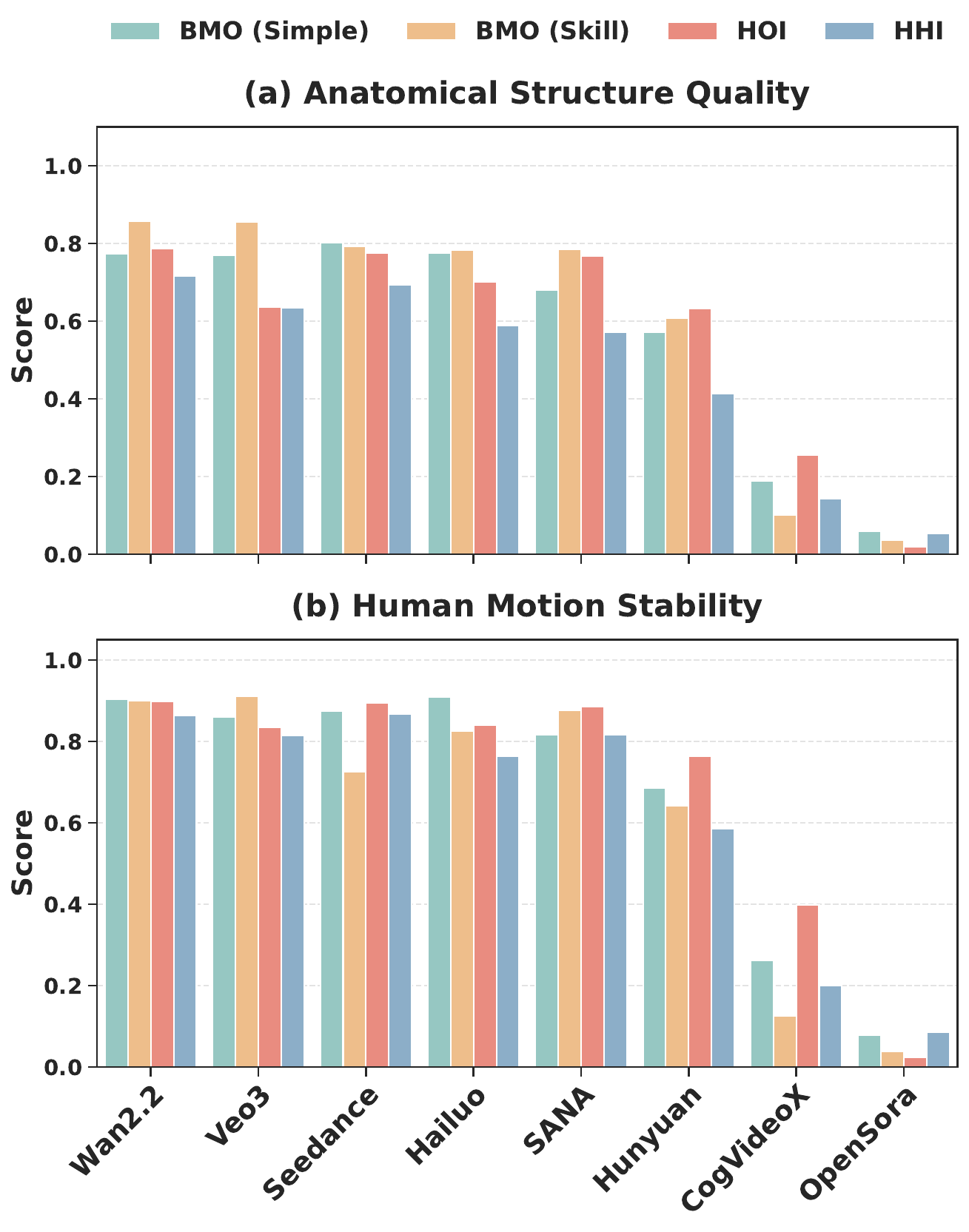}
    
    \vspace{-2mm} 
    
    \caption{Fine-grained Performance Comparison.
    We report the Human Appearance Quality ($S_{Anat}$) and Human Motion Smoothness ($S_{Mot}$) across four distinct prompt categories. }
    \label{fig:category_analysis}
    
    \vspace{-8mm} 
\end{figure}

Figure~\ref{fig:category_analysis} details the model performance across different prompt categories. 
We classify the 100 evaluation prompts into distinct groups, refining the original Body-Motion Only (BMO) category into BMO-Simple (basic daily actions) and BMO-Skill (complex athletic maneuvers). Across these categories, Wan2.2-14B demonstrates superior robustness, maintaining high scores in both anatomical structure quality and motion stability. In contrast, while Veo3 excels in motion generation, it suffers a drop in appearance quality during interaction tasks (HOI and HHI). 
This demonstrates that maintaining anatomical correctness remains a challenge in interaction-rich contexts compared to basic solo-motion tasks.

\subsection{Case study}
\label{sec:case_study}

We analyze two representative cases in Figure~\ref{fig:teaser_right} to highlight the limitations of existing metrics compared to our framework. The top panel shows a man in the kitchen with high-fidelity textures but suffers from a severe anatomical error (specifically, a bifurcated or additional arm). Rule-based metrics like EvalCrafter ($96.77$) and VBench ($1.00$) are misled by the realistic lighting and resolution, assigning high scores despite the structural error. In contrast, our HuM-Eval framework successfully detects the anatomical distortion failure, yielding a low score ($0.33$) that accurately reflects the anatomical error. The bottom panel features a gymnastics video with a physically impossible $180^\circ$ instant flip. VBench ($0.99$) fails to detect this because pixel-level optical flow remains smooth. Surprisingly, the large multimodal model Qwen3-VL also prefers the flipping video ($0.85$), revealing a ``geometric blindness.'' Our framework, however, explicitly tracks 3D global orientation, correctly identifying the sudden rotational jump as a severe artifact ($0.45$).

\section{Conclusion}

In this paper, we introduce HuM-Eval, a human-centric evaluation framework with a coarse-to-fine strategy. Specifically, we combine a coarse-grained Perceptual Visual Prior with fine-grained scores for Anatomical Structure and Motion Stability. Extensive experiments demonstrate that HuM-Eval achieves superior alignment with human preference compared to existing baselines. Building on HuM-Eval, we introduced HuM-Bench to benchmark existing video generation models. We believe these contributions provide a pivotal reference to guide future advancements in human motion generation.

\section*{Acknowledgements}
This work is supported by the National Natural Science Foundation of China (No. 62276268), and SentiPulse Technology Co.

\bibliographystyle{IEEEbib}
\bibliography{icme2026references}

\vspace{12pt}
% \color{red}
% IEEE conference templates contain guidance text for composing and formatting conference papers. Please ensure that all template text is removed from your conference paper prior to submission to the conference. Failure to remove the template text from your paper may result in your paper not being published.

\end{document}